# Human-in-the-Loop Disinformation Detection: Stance, Sentiment, or Something Else?

Alexander Michael Daniel
*Center for Operations Research and Analysis*
*Defence Research and Development Canada*
*Department of National Defence*
Room 119, Building T86, 3701 Carling Avenue
Ottawa, Ontario, Canada,
K2K 2Y7
Alexander.Daniel@forces.gc.ca

# Human-in-the-Loop Disinformation Detection: Stance, Sentiment, or Something Else?[1]

Alexander Michael Daniel

*Defence Research and Development Canada – Center for Operations Research and Analysis*

Alexander.daniel@forces.gc.ca

## Abstract

Both politics and pandemics have recently provided ample motivation for the development of machine learning-enabled disinformation (or "fake news") detection algorithms. Existing literature has focused primarily on the fully-automated case, but the resulting techniques cannot reliably detect disinformation on the varied topics, sources, and time scales required for military applications. By leveraging an already-available analyst as a human-in-the-loop, however, the canonical machine learning techniques of sentiment analysis, aspect-based sentiment analysis, and stance detection become plausible methods to use for a partially-automated disinformation detection system. This paper aims to determine which of these techniques is best suited for this purpose and how each technique might best be used towards this end. Training datasets of the same size and nearly identical neural architectures (a BERT transformer as a word embedder with a single feed-forward layer thereafter) are used for each approach, which are then tested on sentiment- and stance-specific datasets to establish a baseline of how well each method can be used to do the other tasks. Four different datasets relating to COVID-19 disinformation are used to test the ability of each technique to detect disinformation on a topic that did not appear in the training data set. Quantitative and qualitative results from these tests are then used to provide insight into how best to employ these techniques in practice.

## 1 INTRODUCTION

The *information environment*[2] has become an increasingly important area of focus for military operations [1], has become a source of an increasing threat to democratic countries in the form of *disinformation*, i.e. "intentional falsehoods" that are spread through a variety of means to "advance political goals" [3].[3] The significance of this threat has motivated a substantial quantity of research in the field of machine learning and artificial intelligence on methods for the general detection of disinformation (see e.g. overviews in [5-10]).

Given the motivation for the application to defence contexts, this paper is concerned with the detection of disinformation under the following five constraints. First, *the targets and/or substance of disinformation of interest to the user may change over time scales as short as a few days (or even less)*. Not all disinformation will be relevant to a given defence user, and which narratives are of interest can vary day-by-day as mission objectives change and as new narratives arise: for example, over the course of a single weekend in March 2020, the spread of a video of Canadian Armed Forces (CAF) vehicle transportation on social media caused fears about military involvement in Canada's COVID-19 pandemic response [11]. Second, *the disinformation narrative may target novel entities or contain novel substance when compared to previous narratives*. For instance, the COVID-19 pandemic has brought to prominence public health officials like Anthony Fauci who are then subject to disinformation narratives (see e.g. [12]), while for others, like Bill Gates, it has provided new fodder for conspiracies [13]. Third, *the targets and or substance of disinformation narratives may be of little interest to the broader public*. While some narratives receive attention or coverage in popular media (e.g. as in [11]), not all do or will. Fourth, *the domains on which the disinformation is created and/or propagated*

---

[1] © Her Majesty the Queen in Right of Canada, Department of National Defence, 2021

[2] As cited in [1], the Department of Defense defines the information environment as the "aggregate of individuals, organizations and systems that collect, process, disseminate, or act on information" [2].

[3] *Disinformation* is distinguished from *misinformation* in that the latter is the "inadvertent or unintentional spread of inaccurate information without malicious intent" [4] (as cited in [3]), although the distinction is not important to this work. While the term "fake news" is sometime used for either (or both), its use is discouraged in [3], and following their lead, the term "disinformation" is used exclusively in the rest of this paper as a blanket term covering all three.

*may vary in kind*. A disinformation narrative may be propagated in longer-form, infrequently-shared, formal news articles, or single-sentence, widely-shared, informal social media posts, or a variety of media kinds in-between.[4] The *age* of the creator/disseminator may also vary: this could be a minutes-old Twitter account, or a decades-old news media organization. Fifth and finally, *the user of the system should be able to obfuscate their activity,* i.e. do their work without the risk of revealing to adversaries and/or the public their activity.

These constraints jointly entail a set of desiderata for a disinformation detection system, i.e. that the system detects disinformation satisfying each of the constraints. However, these desiderata preclude, to the best of our knowledge, the use of essentially all prominent machine learning-enabled approaches to disinformation detection in the extant literature. Some methods assess the veracity of a claim by explicitly comparing it to the contents of an existing, trusted knowledge base, e.g. [15-19], where the knowledge base can be derived from, for instance, a corpus of scientific abstracts [15], an external fact-checking website [16], or even Wikipedia [17]. In general, these methods will struggle under the first three constraints, as the establishment (or existence) of a knowledge base with the required breadth (for the second constraint) on a topic that may generate little interest to the broader public (as per the third constraint) under the imposed time limit (in the first constraint) is unlikely. Other methods use algorithms to assess the veracity of a claim after training on a dataset of claims that are labelled as being either true or false (or some other finer-grained classification of truth), such that the resulting algorithm represents the knowledge of the dataset implicitly in its parameters. Examples of such datasets include the popular LIAR dataset [20] or the collection of true and false claims about COVID-19 [21] used in the CONSTRAINT challenge at the AAAI2021 conference [22]. These methods will struggle with the same constraints as the knowledge base methods: the algorithm will be ill-equipped to classify claims dealing with knowledge not contained in its training data, and the ability to quickly acquire and prepare new training data is unlikely.

Other disinformation detection methods avoid directly assessing the veracity of the claim made in the text. Some methods evaluate the trustworthiness of the information source [23-25], but these will not have the ability to adjudicate on newer domains (like new social media profiles) as required under the fourth constraint. Diffusion-based methods attempt to separate disinformation from information by identifying differences in how the two types of media spread across the internet [26, 27], but this typically pertains only to social media, and thus runs against the fourth constraint. Organic individual reactions to content can also provide a signal indicating disinformation [28-31], but this only pertains to content on domains where commenting can occur, and since not all domains will have the requisite comment sections, this approach lacks the generality required by the fourth constraint. Finally, other methods seek to exploit a perceived difference in writing styles between reliable information and disinformation [18, 32, 33]; but by the fourth constraint, there is no guarantee that this difference will persist across all domains where disinformation can be found.

With only two exceptions (to be discussed below), these methods appear, to the best of our knowledge, through direct exploration of the literature and the surveys [5-10], to exhaust all prominent *kinds* of approaches to disinformation detection in the available literature (though there are many more *instances* of these kinds not cited here). By the constraints outlined above, these kinds of approaches are inadequate for the envisioned defence application of disinformation detection.

In contrast, this paper discusses methods for the development of a disinformation detection system that does satisfy the aforementioned constraints by exploiting a resource unavailable (in general) to the developers of the aforementioned approaches: an *active* human-in-the-loop.[5] Whereas these previous approaches endeavour to

---

[4] Two other relevant *kinds* not addressed in this work are the language of the disinformation, and the media type, i.e. text, audio, video, image, etc. The focus here and in most of the referenced works is on English text exclusively. For other text in other languages, some of the approaches here will still apply, but will be contingent on, for instance, the availability of adequate datasets. For other media types (regardless of language), detection methods will be of a different kind entirely: see e.g. the related detection problem in the "Hateful Meme Challenge," which requires joint analysis of an image and some text to determine if a meme contains hate speech [14].

[5] Though the specific approach in this paper is novel, the idea of using active human participants in disinformation detection is not (this is one of the two aforementioned exceptions about kinds of approaches to disinformation detection). Works like [19, 34-36] explore the possibility of, broadly speaking, *soliciting* feedback from users (rather than observing the feedback that users have already given, as in [28-31]) in the disinformation detection process. In general, these approaches have trouble with some or all of the third, fourth, and fifth constraints: either the users in question won't know the truth for the content this paper is interested in, the domains won't admit the solicitation

*ICCRTS 2021* 3

automate the entire disinformation detection process for their envisioned applications, in a defence context, the goal is not merely to detect disinformation, but to analyze and understand it as part of one or more larger goals, like (for instance), an analyst maintaining situational awareness of the information environment. Though machine learning models have demonstrated rapid increases in sophistication in recent years, they do not yet appear capable of what might reasonably be labelled "understanding" (see e.g. OpenAI's GPT models and the criticisms thereof by Gary Marcus [37-40]). The approach in this paper is therefore not to attempt to replace the human analyst with a completely automated disinformation detection algorithm, but to augment the human's abilities with machine intelligence, thereby "creating" a human-AI hybrid system that leverages the individual strengths of each entity.

How could such a hybrid system satisfy the aforementioned constraints? One possible way is as follows: For the first constraint, the human user whose disinformation interests may change day-to-day is now part of the system doing the detection, removing the burden on the algorithm to "know" both what is relevant and what is fact versus fiction, on that narrow time scale. To be clear, we take the human user to know the nominal truth regarding the disinformation they're interested in: if, e.g., a defence analyst is searching for sources of disinformation relating to the possibility that the armed forces might be deployed to enforce martial law using COVID-19 as a pretence (as covered in the [11]), the user would be required to know only that the armed forces are indeed *not* going to do that.

What remains for the algorithm to do, then, is to identify relevant disinformation as being that which has the appropriate "approval" or "disapproval" relationship (to be clarified below) to the user-provided topic or claim. To satisfy the second constraint, the algorithm must be able to identify the relevant text regardless of the topic or claim the user is interested in, i.e. it must be an *open-domain* [41] algorithm. Since the algorithm identifies disinformation by comparing the candidate text to the user-provided truth, the third constraint is easily satisified, as no external resources (from, say, public fact-checkers or through crowd sourcing) are required. Moreover, the fourth constraint is also satisfied without issue as the disinformation detection procedure relies only on the text of the purported disinformation and not on information about who created or spread the text. It can thus work on any text-based source, regardless of age or media type. Finally, since the algorithms can be run offline, the obfuscation condition can be satisfied as long as text to be analyzed is obtained (or "scraped") from the internet in a manner consistent with the fifth condition (though discussion of this is beyond the scope of this paper).

In summary, the human-AI hybrid disinformation detection system proposed in this work functions as follows: The human user identifies a topic or claim about which they believe disinformation will be found, and seeks to learn more about what is being said and (possibly) who is saying it. The machine learning algorithm, taking in the topic/claim and large amounts of text scraped from the internet, is able to, by using the appropriate approval/disapproval relationships (as decided by the user), separate the text into two groups: relevant disinformation, and everything else (including both information, and disinformation that doesn't pertain to the topic of interest). Closing the loop, upon seeing the results of the search, the human can determine if it satisfies their interests or not, and if necessary, revise the search and run the process again.

The remainder of this paper, then, is dedicated to examining the relative merits of three plausible candidates for this algorithm. The first is sentiment analysis, which seeks to capture and characterize the sentiment of the input text as a whole. The second is aspect-based sentiment analysis, in which each sentiment-arousing object (i.e. "aspect") in a text excerpt is identified, along with the sentiment it aroused. Finally, stance detection algorithms seek to elucidate one excerpt of text as being for, against, or unrelated to another excerpt of text. The idea of using stance detection for disinformation is not itself new (this is the second exception referred to above). The first "Fake News Challenge" (FNC-1) [42] used a dataset that takes pairs of news headlines and news stories and labels them with "agree" (i.e. the headline agrees with the text of the story), "disagree," "discuss" (for headline/story pairs that are related but that neither agree nor disagree), and "unrelated." An algorithm trained on this dataset is not, however, a complete disinformation detection system. In the parlance of this report, it runs afoul of the first constraint outlined above - even in the ideal case, it merely knows what agrees, disagrees, discusses or is unrelated to any given input sentence, but does not have a mechanism for determining what that sentence should

---

of user feedback, the request for feedback would risk revealing the defence client's operational interests or some combination thereof.

be, as noted in [43]. For more discussion on stance detection and disinformation, see the recent survey [44].

The purpose of this paper is to try and determine how well each kind of algorithm can be used for disinformation detection in the kind of hybrid system outlined in this paper. In short, this will be done by taking datasets containing disinformation and information on a topic not seen during model training, COVID-19, and seeing how well each model picks out the disinformation relevant to pre-specified topics. In Section 2 further background on the literature surrounding these topics is provided, and informs Section 3, wherein the model architecture and training are discussed. In Section 4, the models are tested, and in Section 5, the test results are are analyzed as a whole, with noteworthy trends identified and a path for future research outlined.

## 2 RELATED WORK

### 2.1 DISINFORMATION DETECTION

One final caveat on the disinformation detection literature: the work in this paper is different from works like [45-47], which are broadly concerned with the titular question of [45], namely "Which Machine Learning Paradigm for Fake News Detection?". This may sound like the question posed in the title of this paper, but the difference is that those papers are concerned with deciding which learning algorithm *architectures* (trained on the same training dataset) perform best when tested on a test dataset extracted from the same source as the training dataset. In this paper, however, we will use near-identical architectures (see Section 3.2) trained on distinct datasets (and thus representing distinct *tasks*), and tested on a variety of test datasets from a task different from any of the training datasets, to investigate which *training task* is best suited for use in the test task.

### 2.2 TRADITIONAL SENTIMENT ANALYSIS

By "traditional" sentiment analysis, we broadly mean the task of assigning a single sentiment to the entirety of some input text, which may include one or more sentences. Binary sentiment analysis tasks require the input text to be labelled as positive or negative, while finer-grained sentiment analysis tasks may allow for a neutral label (i.e. no sentiment), or degrees of positive/negative sentiment (e.g. "very positive"). So construed, traditional sentiment analysis has been the subject of an enormous amount of study in the literature, with popular datasets like the IMDB movie review dataset [48] and Stanford Sentiment Treebank (SST) dataset [49] having been used in thousands of papers in the years since they were first introduced (with roughly 2500 and 4900 citations, respectively, according to Google Scholar at time of writing). Performance on binary sentiment analysis now approaches human level: the leaderboard for the General Language Understanding Evaluation (GLUE) benchmark [50] reports that the top-performing algorithm on the benchmark now achieves a human-level accuracy (97.8%) on the SST dataset.[6] For our purposes, it is sufficient to note that Google's large transformer language model BERT, with only a single softmax layer afterward, achieves near state-of-the-art performance on the binary SST dataset, with the "base" version of BERT yielding 93.5% accuracy [51]. BERT is also used for near state-of-the-art performance in both aspect-based sentiment analysis and stance detection (see Sections 2.3 and 2.4), and will thus form the backbone of the architectures used in this paper (see Section 3.2).

### 2.3 ASPECT-BASED SENTIMENT ANALYSIS

In contrast with traditional sentiment analysis, aspect-based sentiment analysis is concerned with identifying both sentiment and the targets thereof, potentially allowing for sentiments about multiple targets to be classified within the same sentence. Sentences like "The laptop has nice screen resolution, but the boot-up speed is slow" express both positive and negative sentiments about an entity (i.e. the laptop), and therefore it would be inappropriate to label the sentence with a single sentiment, as would be done by a traditional sentiment analysis algorithm. Note also that "aspect-based sentiment analysis" is sometimes used in the literature to refer to distinct but related tasks (see e.g. differences in the surveys [52] and [53]). This paper is concerned with "aspect-based sentiment analysis" in the sense of what the popular SemEval challenges (2014 Task 4 [54] and 2015 Task 12 [55]) call "aspect term" sentiment classification, in which the target of the sentiment must be a noun phrase appearing in the sentence. This is also sometimes called "targeted" sentiment analysis, as in [41, 56]. More recent works in this line of research exploit powerful transformer models like BERT, as in [57-61]. Perhaps the most popular datasets for this task are the ones derived from the SemEval aspect-based sentiment analysis challenges in 2014 [54], 2015 [55], and 2016 [62], which consist of reviews of laptops and restaurants. Top performers on these datasets [57-61] now yield F1 scores in the low to mid-sixties on the laptop dataset, while the restaurant dataset yields scores in the mid-seventies.[7] Though these approaches sometimes differ in specific architecture, for the purposes of this work, we simply note that near-state-of-the-art models can be obtained using BERT as a word embedder, and with only a few task-specific layers added on top. More details on the specific BERT-based architectures used here are given in Section 3.2.

---

[6] See the leaderboard at gluebenchmark.com/leaderboard

## 2.4 STANCE DETECTION

As with aspect-based sentiment analysis, "stance detection" encompasses a number of related but distinct methods and/or tasks relating to the characterization of the agreement (or lack thereof) that different bodies of text have to one another. In the popular dataset from the SemEval 2016 Task 6 challenge [68, 69], algorithms are given a topic in the form of a noun phrase (e.g. "Hillary Clinton" or "Donald Trump" [68]) and some input text, and are required to determine if the text is for or against the topic (or neither). This requires the algorithm to learn contextual information about the topics of interest, a fact which precludes this kind of approach from use in disinformation detection as conceived of here as it violates the first two constraints outlined in Section 1. Noting this limitation, "cross-target" sentiment detection [70] is a variant on stance detection that aims to develop algorithms using data from one domain for use on domains unseen during training. However, these unseen domains must be known in advance, i.e. there are specific "target" domains for the algorithm to apply to, as in [70-73] and subsequent work, so these are not the fully open-domain approaches needed for the disinformation detection task of this paper. Work on improving algorithm performance on unseen (during training) and unknown (prior to testing) topics for this dataset is, to the best of our knowledge, limited to [74].

However, the FNC-1 dataset [42] is far better-suited to training algorithms for open-domain use, as the articles from which the data is derived cover a wide array of topics and claims, and the task only requires the algorithm to determine the relative stance of the two claims (though on the downside, such an algorithm would not learn the contextual knowledge required that could be used to infer stance as discussed above). While the top-performing algorithm on this dataset achieves a 97.8% accuracy, near-state-of-the-art algorithms with accuracies in the low-to-mid nineties can be achieve using BERT as a word embedder with minimal task-specific architecture thereafter [75, 76] (though the latter technically uses the BERT variant RoBERTa [77]).

It should be noted that the FNC-1 dataset has also received critical attention: [78] discusses the extreme class imbalance in the dataset, with the "unrelated" label constituting 72.8% of labels, while the "disagree" label applying only to 2.0% of headline/article pairs. As a result, the classifiers they assessed show strong performance distinguishing unrelated pairs from related ones (most achieved over 0.99 F1-score at this phase), but showed much poorer performance identifying agreeing and disagreeing pairs. In studying this problem, they analyze successful features, propose an alternative evaluation metric that accounts for the class imbalance, and modify an additional dataset [79] for use on stance detection in order to improve the agree/disagree performance. As the distinction between the "agree" and "disagree" classes is crucial to the application of stance detection to disinformation detection, these issues will be revisited when developing the training dataset for stance detection in Section 3.3.3.

## 2.5 SOMETHING ELSE?

Finally, we note that stance and sentiment do not exhaust the possible approaches to this disinformation detection task provided by canonical machine learning tasks. For instance, auxiliary sentence construction [80], question answering (QA) methods [81, 82, 83], and natural language inference (NLI) methods [84] have been explored as means for improving performance in aspect-based sentiment analysis and stance detection, and as such, might be plausibly employed for disinformation detection as construed in this paper. It is however, beyond the scope of this paper to explore that possibility.

## 3 MODEL PREPARATION

### 3.1 APPLYING STANCE AND SENTIMENT TO DISINFORMATION DETECTION

Having discussed the three tasks and their relevant literature in more detail, we are now in a position to explain how they are to be used in a human-AI hybrid system capable of disinformation detection tasks of interest in this paper. For convenience, the three techniques will hereafter be referred to by their initials: "SA" for (traditional) sentiment analysis, "ABSA" for aspect-based sentiment analysis, and "SD" for stance detection.

---

[7] The recent paper [63] achieves even higher scores, but uses a domain-specific pre-training method that make that approach ineligible to the application in this paper (as per the rst and second constraints in Section 1). When the pre-training method is not used, the resulting performance is more in line with the aforementioned works. Other noteworthy recent works in aspect-based sentiment analysis [64-67] are likewise ineligible for further consideration here because they consider variations on the aspect-based sentiment analysis problem that are not relevant to their use for disinformation detection.

First, the human must know/decide what topic (paradigmatically a noun phrase like "Canada" or "Bill Gates") or claim (paradigmatically a sentence like "The 2020 US Election was stolen from Donald Trump" - see e.g. [85] for discussion of this conspiracy theory) that they are interested in detecting disinformation about. For SA and ABSA, keywords representing the topic or claim must be chosen, though this will be easier for topics than for claims: using the examples above, sets of keywords could be "Canada", "Bill Gates", or "2020 Election", "stolen", and "Trump". Then disinformation is identified by picking a sentiment one expects to be associated with the topic or claim e.g. "negative" about "Bill Gates". Then input text is considered disinformation if it contains the keywords and has the identified sentiment. A possible weakness for the use of SA immediately arises – it's not necessarily the case that all negative text mentioning Bill Gates is disinformation; indeed, it remains to be seen if such approaches result in to many false positives.

The ABSA approach provides a plausible remedy to this weakness. The user is able to decide which keywords are the targets of which emotions (though not all keywords necessarily need emotions). Using the above example, one might use ABSA to search for text that is negative about "US Election" but positive about "Trump." This could help narrow down results from all those that may merely be discussing the topic with a negative sentiment but not supporting the conspiracy theory itself.

For SD, claims are easy to accommodate: take the claim (or some variation thereof) as the auxialliary text for the algorithm. For broader topics, the user would devise a broad claim to use as auxiliary text, e.g. "Bill Gates" might work by itself, but "Bill Gates is the worst" or "I hate Bill Gates" might work better at catching disinformation about Bill Gates. Different possibilities will be explored in Section 4. A possible strength of the stance detection approach is that it won't necessarily need the appearance of specific words (i.e. the keywords used for SA and ABSA) to understand agreement/disagreement, e.g. it may be able to understand "I dislike Bill Gates" as agreeing with "I hate Bill Gates" despite the absence of the word "hate" in the former text.

### 3.2 MODEL ARCHITECTURES

In order to facilitate as-fair-as-possible comparison between the efficacy of the three algorithms, the model architectures used for each should be as similar as possible. Fortunately, as discussed in Section 2, near state-of-the-art results in each task have been achieved by simple architectures using (the "base" version of) BERT as a word embedder with minimal architecture-specific layers added thereafter. This provides a roughly equal playing field in which to assess the relative performance of each approach.

Following the standard procedure for BERT [51], for each task, the input text (or in the case of stance detection, the claim and input text) are tokenized into Wordpiece embeddings, with the special tokens "[CLS]" and "[SEP]" added to the front and back of the text respectively (and for stance detection, in between the claim/auxiliary text and input text). Due to limitations of GPU size, a maximum of 192 tokens (including the special tokens) were used as input, with shorter input sequences zero-padded and longer sequences truncated. The output from BERT is a vector of length 768 for each input token, for a total of 192x768 = 147,456 output values. For SA, these values were fed into a single three-node soft layer corresponding to "positive," "neutral," and "negative" outputs. For SD, the BERT output was fed into a four-node softmax output layer corresponding to the "unrelated," "agree," "discuss" (related but neutral), "disagree," classes of [42]. Finally, for ABSA, each token requires its own four-node softmax output layer, with outputs corresponding to "O" (the null token), "positive," "neutral," and "negative." Although an architecture where all 192 groups of 4 nodes were allowed access to all 147,456 output values was explored, it was found through informal experimentation that restricting the input for a single 4 node token output to only the length-768 vector corresponding to that token actually improved performance, despite the restriction of the feasible space, seemingly preventing the optimizer from being stuck in one of the many worse local minima available in the larger feasible space. The architecture ultimately used for the ABSA model tested in this paper feeds each token's 768-long vector into a 4-node softmax output layer, where the softmax layer for each token is held to have identical parameters to the other softmax output layers, i.e. in addition to the parameters for BERT, the ABSA model has only an additional 4(768 + 1) = 3076 parameters to be trained in the output layer.

### 3.3 TRAINING DATA

For the sake of fair comparison between the tasks, the training dataset used for each model needs to be the same size. The ABSA task was the most challenging to find data for; as a result, the training datasets were limited to 25,000 items for training, with approximately 2,000 items for validation. Note that all text, prior to the Wordpiece tokenization discussed above, was "cleaned" to remove hyperlinks, emojis, and other special characters. Notably, "hashtags" from tweets (contiguous strings of characters following the # character, e.g. "#Jeb2016" in the example from Section 2.4) were not removed or modified, as they were sometimes the only source of information about the target of sentiment of a tweet. Twitter handles were likewise not removed. All text was converted to lowercase form before being processed by the algorithms.

### 3.3.1 Sentiment Analysis

First, all 11,855 sentences in the Stanford Sentiment Treebank dataset [49] were used,[8] with the "very positive" and "positive" classes collapsed to a single "positive" label, and "very negative" and "negative" similarly collapsed into one "negative" class. To fill out the remaining 13,145 items required, the Yelp and Amazon review datasets from [86],[9] containing sentences along with their one to five star ratings, were used. Like with the SST dataset, we counted one and two star ratings as "negative," three star ratings as "neutral," and four and five star ratings as "positive." From the Amazon test dataset, the first 1676 positive, 3054 neutral, and 1842 negative reviews were taken, and with identical numbers (except for one additional positive example) from the Yelp test dataset taken in the same fashion. The result, given that the SST dataset has 4980 positive, 2226 neutral, and 4649 negative examples, is a dataset of 25,000 items with balanced classes of 8,333 items each (plus one extra for the neutral class). To generate the validation dataset, the first 333 positive, 333 negative, and 334 neutral reviews from the Amazon training dataset were taken, with a similar balance taken from the Yelp training dataset, to obtain an approximately balanced validation dataset with 2,000 items in it.

### 3.3.2 Aspect-Based Sentiment Analysis

The training and validation datasets used for ABSA are the result of combining six different datasets. Unlike the SA and SD datasets, the resulting datasets aren't balanced: first, since most words in the sentence are labelled with the null token, that null class dominates the other three, and second, the relative scarcity of data meant that to get the dataset as large as possible, considerations about having equal numbers of positive, negative, and neutral examples were sidelined.

The first dataset used was the aforementioned SemEval2014 Task 4 dataset [54];[10] between the "laptop" and "restaurant" training and test datasets, 9,880 examples were obtained. The next dataset, from [87],[11] was obtained by taking a subset of sentences from the SemEval2014 Task 4 datasets and transforming them using negation and "speculation," where, for instance "I like the food" becomes "I don't like the food" (negation) and "I'm not sure if I like the food" (speculation), with the labels modified accordingly. This added an additional 2423 items to the dataset. A second related dataset, the "multi-aspect multi-sentiment" (MAMS) dataset [88],[12] takes restaurant reviews from the same basic corpus as the SemEval2014 Task 4 restaurant, with the caveat that each review must contain more than one target and kind of sentiment, adding another 5297 items to the dataset. Another 2358 pieces of data came from a Twitter dataset used in earlier work on targeted sentiment analysis [41].[13] The YASO ("Yelp, Amazon, SST, Opinosis") dataset [89] contains data from the four named SA datasets and relabels them for ABSA; this resulted in another 1989 examples.[14] Finally, the "Sentihood" dataset [90] yields 5215 items discussing sentiments about neighbourhoods in London, England.[15]

---

[8] Available at nlp.stanford.edu/sentiment
[9] Available at github.com/zhangxiangxiao/Crepe/tree/master/data
[10] Available at https://alt.qcri.org/semeval2014/task4/index.php?id=data-and-tools, though see also https://github.com/huminghao16/SpanABSA from [59].
[11] Available at https://github.com/jerbarnes/multitask_negation_for_targeted_sentiment
[12] Available at github.com/siat-nlp/MAMS-for-ABSA
[13] Available at m-mitchell.com/code, though see also https://github.com/huminghao16/SpanABSA from [59].
[14] Available at www.research.ibm.com/haifa/dept/vst/debating_data.shtml#TargetedSentimentAnalysis. This link does not actually host the text itself, but rather contains sentence annotations and a procedure for recovering the text plus annotation pairs once one has already downloaded the four datasets above. From the 2015 total annotations from the four aforementioned datasets, 1989 were successfully recovered, with a few examples from each dataset yielding errors, a difference we take to be negligible in practice.
[15] The link to the dataset provided in the original paper is now dead, but the data can be obtained from other sources, e.g. github.com/Nix07/Utilizing-BERT-for-Aspect-Based-Sentiment-Analysis, which is what was used for this paper. The dataset has the actual location names removed and replaced with placeholders (like "LOCATION1" and "LOCATION2"), so to function in practice, the placeholders must be replaced with real nouns. This was achieved using the "Canadian Geographical Names" dataset (from Natural Resources Canada, available at open.canada.ca/data/en/dataset/e27c6eba-3c5d-

Combined, these six dataset yielded 27,162 items with which to train and validate the ABSA algorithm. Taking 25,000 for training, the remaining 2,162 were used for validation (specifically, the full Sentihood test set and the first Sentihood validation set, less the first 76 items).

### 3.3.3 Stance Detection

The obvious choice for SD data is the aforementioned FNC-1 dataset [42].[16] As discussed in Section 2.4, class imbalances in the dataset can result in poor performance on the important "agree" and "disagree" classes. In order to develop a balanced training dataset with 25,000 examples, we took from the FNC-1 training data all 840 "disagree" pairs, all 3,678 "agree" pairs, and the first 5,346 "discuss" and 3,125 "unrelated" pairs. To supplement these, two additional stance datasets were used. The first is the "Argument Reasoning Comprehension" (ARC) Task dataset [79], as used and discussed in [78],[17] from which we took all 1,257 "agree" pairs, all 1,402 "disagree" pairs, all 904 "discuss" pairs, and the first 3,125 "unrelated" pairs. The second is the "Perspectrum" dataset [91],[18] from which we took an additional 1315 "agree" and 4008 "disagree" pairs. In sum, this yields 6,250 examples from each class. For a validation set, the first 500 examples of each class were taken from the FNC-1 competition test dataset.

### 3.4 MODEL TRAINING AND VALIDATION

Each model was trained using the Adam [92] optimization algorithm with a learning rate of $10^{-5}$. The SD and SA algorithms were trained using only a single epoch, while the ABSA algorithm was trained for six epochs, as these were found to yield the minimal training set error without overfitting on the validation set. For each model, the training method was run 5 times on the final hyperparameter set, and the model with the highest categorical accuracy on the validation dataset was used for subsequent testing (Section 4).

The SA algorithm obtained a 71.35% categorical accuracy on its validation dataset, while the ABSA algorithm obtained a 99.47% score in the same metric. This latter value is somewhat less informative though, as the dataset is dominated by easily identified null tags. More meaningful metrics for ABSA are the "exact-match" (entity-level) precision, recall, and F1 (see e.g. [59]), which capture how accurately the algorithm identifies both the sentiment and boundaries of the target in the sentence. For the sake of future comparison, exact-match precision, recall, and F1 scores of 0.612, 0.695, and 0.651 were achieved on the validation dataset respectively. Finally, the top-performing model on the stance detection validation dataset achieved a categorical accuracy of 70.45%. Though this may seem low compared to the low-to-mid-nineties scores discussed in Section 2.4, it is due to the difference in class balance of the two datasets. When tested on the full FNC-1 competition test set, the algorithm achieves a 90.8% categorical accuracy.

## 4 MODEL TESTING AND RESULTS

To the best of our knowledge, there are no datasets available in the literature that have the exact features to directly compare the three models against one another; ideally, such a dataset would contain labels for SD, SA, and ABDA for intra-task evaluation (i.e. seeing how algorithm performance drops compared to training/validation results), along with true/false labels and labels denoting topic and/or claim so that text algorithms that correctly return large amounts of false information do not get credit for returning false information unrelated to the claim or topic being searched for. The tests in this paper were therefore necessarily indirect, but taken together, were sufficient to reveal some insight about the utility of the various algorithms for the task of human-in-the-loop disinformation detection. There are three kinds of tests run, to be elaborated upon respectively in the next three subsections.

### 4.1 INTRA- AND INTER-TASK TESTING

The intra- and inter-task tests do two things: first, they test each algorithm on a dataset with the same kind of labels it was trained on, in order to establish a baseline for performance for each algorithm on a dataset from a source not used during training or validation, and second, they establish a baseline for how well each algorithm can

---

4051-9db2-082dc6411c2c, or more specifically, the file "cgn_canada_csv_eng.zip" available at the English CSV dataset link on that page), a dataset of Canadian geographic object names, i.e. names of cities, islands, mountains, rivers, and so on. These names were selected from uniformly at random to replace the placeholders. This resulted in some factually false sentences (an actual example: "And Lac Boomerang is ten mins direct on the the tube to Shoal Lake," but these locations are, in reality, separated by approximately two thousand kilometres.), but this is not an issue, as the point of using this data is not to teach the algorithm to learn facts about Canadian geography, but to learn how the relationships between words in a sentence relate to the sentence's sentiment and targets thereof.

[16] Available at https://github.com/FakeNewsChallenge/fnc-1
[17] The dataset is available from the link provided in [78], github.com/UKPLab/coling2018_fake-news-challenge.
[18] Available at github.com/cogcomp/perspectrum

be used to do the tasks of the other algorithms. The latter is important to help determine why the various algorithms perform the way they do on the disinformation detection tasks in subsequent sections. If, for instance, the sentiment algorithms yield poor performance on the stance dataset of Section 4.1.1, then poor performance on the stance-based disinformation detection tests of Section 4.2 is to be expected; however, if those algorithms perform well in Section 4.1.1 but poorly in 4.2, this could suggest that they are less suitable for disinformation detection.

### 4.1.1   Stance Detection and Sentiment Analysis

The test for both SD and SA uses the dataset from [93, 94],[19] which takes the SemEval2016 Task 6 stance detection dataset [68, 69] and annotates it for sentential sentiment as well (with positive/negative/neutral labelling). This test uses the subsets of the data (from both the nominal training and test sets of the original splits in [93]) that focuses on Hillary Clinton and Donald Trump, separated into two distinct test sets (one for each politician). For each politician, two tests were run: one trying to identify favourable stances, and one trying to identify negative stances. Note that in this dataset, text can be for or against a target without explicitly mentioning that target. Nevertheless, for SA, we took input text to favour a candidate if it contained their first name, their last name, or their twitter handle (because the data are taken from tweets), and the overall sentence sentiment is positive. For ABSA, the same keywords were used, but at least one of the keywords had to have a positive sentiment associated with it. The keywords themselves were used as a baseline search, where text was considered favouring a candidate if it contained any of the aforementioned keywords, regardless of sentiment. For stance detection, we tested seven different sentences. Let *f* denote the first name of the candidate, *l* denote the last name, and *t* denote their twitter handle. Then the following sentences were tested: "*f l*", "*t*", "*f l* or *t*", "I like *f l*", "I like *f l* and *t*", "I think *f l* is the best candidate for president of the United States", and "I think *f l* or *t* would be the best candidate for president of the United States". For searches for favourable views towards a candidate, we looked for text the algorithm labelled with the "agree" class (i.e. agreeing with the particular claim above), while for unfavourable views, we looked for text that was labelled with the "disagree" class. Finally, since the data is also labelled for sentiment, we also tracked the overall accuracy of the SA algorithm as a test. The results are shown in Tables 1 and 2; note that the keyword search results are represented by "K" (here and through the rest of the paper).

Finally, the SA accuracy on the Hillary Clinton data was 0.65, and 0.66 on the Donald Trump data – a slight but non-negligible drop in performance from the validation data score of 0.71, likely due to the out-of-distribution test data.

*Table 1: Precision, recall, and F1 scores, for searches of views for and against "Hillary Clinton"*

|      | FOR   |       |      | AGAINST |       |      |
|------|-------|-------|------|---------|-------|------|
|      | Prec. | Rec.  | F1   | Prec.   | Rec.  | F1   |
| K    | 0.28  | 0.90  | 0.43 | 0.71    | 0.64  | 0.67 |
| SA   | 0.53  | 0.64  | 0.58 | 0.88    | 0.32  | 0.47 |
| ABSA | 0.62  | 0.37  | 0.46 | 0.94    | 0.17  | 0.29 |
| SD 1 | 0.31  | 0.66  | 0.42 | 0.67    | 0.53  | 0.59 |
| SD 2 | 0.24  | 0.73  | 0.36 | 0.66    | 0.41  | 0.51 |
| SD 3 | 0.40  | 0.48  | 0.44 | 0.67    | 0.37  | 0.48 |
| SD 4 | 0.32  | 0.69  | 0.44 | 0.68    | 0.54  | 0.60 |
| SD 5 | 0.27  | 0.76  | 0.40 | 0.65    | 0.38  | 0.48 |
| SD 6 | 0.51  | 0.55  | 0.53 | 0.69    | 0.36  | 0.47 |
| SD 7 | 0.46  | 0.56  | 0.51 | 0.69    | 0.41  | 0.51 |

The keyword search recall values show that when the candidates are favoured, the text tends to contain that candidate's name (0.90 for Clinton and 0.99 for Trump), while opposition to the candidate can be expressed without naming them (0.64 recall for "against" Clinton keyword searches, and 0.55 for Trump). The resulting keyword search precision values then simply reveal what proportion of text mentioning the candidate are for/against them. A high recall can be achieved simply by chance (e.g. for "against" Clinton a precision of 0.71 is achieved simply because 71% of text containing her name is text containing attitudes against her), so it is important to demonstrate an even higher precision than the keyword search. The SA approach successfully improved upon the keyword search in all four cases, with the magnitude depending on how successful the keyword search was. In practice, this can be thought of in terms of the human performing a keyword search, and then using sentiment to try and filter out unrelated results. In this case, SA would be successful in narrowing down the results to a pool of more relevant results than the keyword search alone would provide. The cost of this success, however, is a reduced recall - some results that are relevant would be excluded. This effect is more pronounced when using ABSA: in three of four tests, the ABSA algorithm yielded an even higher precision, at the cost of an even worse recall (in all four tests).

---

[19] Available at saifmohammad.com/WebPages/StanceDataset.htm

*Table 2: Precision, recall, and F1 scores, for searches of views for and against "Donald Trump"*

|      | FOR   |       |      | AGAINST |       |      |
|------|-------|-------|------|---------|-------|------|
|      | Prec. | Rec.  | F1   | Prec.   | Rec.  | F1   |
| K    | 0.46  | 0.99  | 0.63 | 0.52    | 0.55  | 0.54 |
| SA   | 0.70  | 0.60  | 0.65 | 0.69    | 0.30  | 0.42 |
| ABSA | 0.66  | 0.30  | 0.41 | 0.76    | 0.19  | 0.30 |
| SD 1 | 0.38  | 0.49  | 0.43 | 0.47    | 0.34  | 0.39 |
| SD 2 | 0.33  | 0.67  | 0.44 | 0.45    | 0.13  | 0.20 |
| SD 3 | 0.62  | 0.46  | 0.53 | 0.49    | 0.29  | 0.46 |
| SD 4 | 0.47  | 0.51  | 0.49 | 0.46    | 0.46  | 0.46 |
| SD 5 | 0.42  | 0.78  | 0.55 | 0.45    | 0.35  | 0.40 |
| SD 6 | 0.54  | 0.28  | 0.37 | 0.44    | 0.26  | 0.32 |
| SD 7 | 0.67  | 0.49  | 0.56 | 0.51    | 0.45  | 0.48 |

The picture for the stance detection algorithm is less rosy. At best, the various queries yield results that are comparable to the basic keyword search, but are more often clearly worse off. Patterns about the best kind of query to use for stance detection are also unclear. The "favour" Clinton search sees the longer, more descriptive claims fare better, but this trend is not at all obviously present in the other three queries.

4.1.2    Aspect-Based Sentiment

The ABSA takes an ABSA dataset and sees how SA and SD will fare compared to an ABSA algorithm when used to do the ABSA task. The dataset, from [95],[20] contains a collection of over 6000 tweets with a 3-valued sentiment (positive, neutral, and negative) labelled for an entity in the sentence. Distinct from the SD dataset used in the last section, this dataset only has text that explicitly names the targets of the sentiment (which, as we will see, will result in a perfect recall score of 1.0 for the keyword search). There are numerous entities who are the subject of sentiment in this dataset. For the purposes of this test we select two: the pop music singer Britney Spears (the target in 924 of the dataset's sentences) and the former American president Jimmy Carter (the target in 101 sentences). For each person, we test one search for positive sentiment and one for negative. The keyword search returns a match whenever the first name or the last name appears in the text, regardless of the target sentiment of the search. The SA approach declares a match when the keywords match as above and if the declared sentiment of the algorithm matches the sentiment of the search. For ABSA, the search is similar, but the algorithm must identify either one of the keywords as having the target sentiment. For SD (letting *n* be the full name of the target), the following three queries are tested: "*n*", "I like *n*", and "I think *n* is the best", with "agree" used as a proxy for "positive" and "disagree" for "negative." The results are shown in Tables 3 and 4.

*Table 3: Precision, recall, and F1 scores, for searches of positive and negative sentiments about "Britney Spears"*

|      | POSITIVE |      |      | NEGATIVE |      |      |
|------|----------|------|------|----------|------|------|
|      | Prec.    | Rec. | F1   | Prec.    | Rec. | F1   |
| K    | 0.29     | 1.0  | 0.45 | 0.20     | 1.0  | 0.33 |
| SA   | 0.56     | 0.88 | 0.68 | 0.48     | 0.72 | 0.57 |
| ABSA | 0.46     | 0.94 | 0.62 | 0.66     | 0.42 | 0.51 |
| SD 1 | 0.13     | 0.84 | 0.22 | 0.09     | 0.65 | 0.16 |
| SD 2 | 0.14     | 0.80 | 0.23 | 0.07     | 0.71 | 0.12 |
| SD 3 | 0.15     | 0.82 | 0.25 | 0.09     | 0.67 | 0.16 |

As expected, the keyword search recall is one, as the target always appears in the sentence about them. The keyword search precision values show that, for Britney Spears, about half of the text about her in the dataset is neutral, while for Jimmy Carter, only roughly one-fifth of the data about him has any sentiment at all.

*Table 4: Precision, recall, and F1 scores, for searches of positive and negative sentiment about "Jimmy Carter"*

|      | POSITIVE |      |      | NEGATIVE |      |      |
|------|----------|------|------|----------|------|------|
|      | Prec.    | Rec. | F1   | Prec.    | Rec. | F1   |
| K    | 0.03     | 1.0  | 0.07 | 0.17     | 1.0  | 0.30 |
| SA   | 0.07     | 0.80 | 0.13 | 0.47     | 0.96 | 0.63 |
| ABSA | 0.08     | 0.60 | 0.14 | 0.48     | 0.64 | 0.55 |
| SD 1 | 0.09     | 0.65 | 0.16 | 0.04     | 0.60 | 0.07 |
| SD 2 | 0.07     | 0.71 | 0.12 | 0.01     | 0.72 | 0.03 |
| SD 3 | 0.09     | 0.67 | 0.16 | 0.02     | 0.56 | 0.03 |

Once again, the SA and ABSA algorithms are able to improve the precision of the keyword search by a factor of two to three, with the ABSA approach further improving, if only narrowly, on the sentential approach in three of the four tests. Despite a reduced recall, the overall F1 improves upon the keyword F1. The stance detection algorithms, however, do not fare so well. While maintaining strong recall values in general, the precisions are very poor: the algorithm identified most of the relevant text, but failed to filter out the large quantity of neutral information about the two targets.

Finally, we note briefly that this was also an opportunity to test the ABSA algorithm on the ABSA task. For three of the tests in this section, the resulting F1 was in the fifties and low sixties, though this is not a direct comparison to the validation data results as the matches in these tests did not have to be exact. Recalling that the exact-match F1 reported in Section 3.4 on the validation data was 0.651, this constitutes a non-negligible drop in performance likely due at least in part to the use of out-of-distribution data, but also likely due in part to the extreme class imbalance - the largest class (neutral) for both targets is the one that each test class (positive and negative respectively) is most likely to be confused for. Indeed, it is plausible to explain the performance in the fourth test (an F1 of 0.14 on the positive Jimmy Carter test) in precisely these terms: with a very small number of positive cases and a large number of neutral cases, even a small rate of confusion of neutral for positive will yield a large number of false positives in absolute terms.

---

[20] Available at goo.gl/5Enpu7

4.2 STANCE-BASED DISINFORMATION DETECTION TESTING

Two datasets fall under this classification. Both contain pairs of claims and text that are labelled as agreeing, disagreeing, and being unrelated (or something equivalent), where the claims themselves are false. These tests attempt to mimic the ability of the human-in-the-loop to find text they believe to be false by identifying text that agrees with a false claim or disagrees with a true claim.

4.2.1 The "COVID-CQ'' Dataset

The COVID-CQ dataset contains tweets along with labels indicating agreement with, disagreement with, or being unrelated to the claim "Cholorquine/Hydroxychloroquine are cures for the novel coronavirus" [96] (see e.g. [97] and [98] for why this counts as disinformation).

For SA, we used a search that will declared disinformation if the sentence contains either "chloroquine" or "hydroxychloroquine" and has positive sentiment. Likewise, the ABSA approach declared disinformation when a sentence contained either of those words, and either one of the words had a positive tag associated with it. For SD, two sets of tests were conducted. The first set of tests will search for agreement with a claim, using the following nine claim variations, with $x_1$ = "hydroxychloroquine", $x_2$ = "chloroquine", $c_1$ = "COVID", and $c_2$ = "the coronavirus":

- "$x_i$", $\forall i \in \{1,2\}$ (2 claims)
- "$x_i$ is a cure for $c_j$", $\forall i,j \in \{1,2\}$ (4 claims)
- "Either $x_1$ or $x_2$ is a cure for $c_j$" $\forall j \in \{1,2\}$ (2 claims)
- "Either $x_1$ or $x_2$ is a cure for $c_1$ or $c_2$" (1 claim)

These nine claims will respectively be labelled 1 to 9, with the four claims of the second item above using the following variable pairs respectively for claims 3 through 6: ($x_1$, $c_1$), ($x_1$, $c_2$), ($x_2$, $c_1$), and ($x_2$, $c_2$). The second set of claims is obtained by adding "It is not the case that" to the beginning of each sentence; these will be referred to as the "negated" claims. Table 5 shows the results.

Table 5: Precision, recall, and F1 scores for a search for disinformation about (hydroxy)chloroquine

|      | REGULAR CLAIMS | | | NEGATED CLAIMS | | |
|------|-------|------|------|-------|------|------|
|      | Prec. | Rec. | F1   | Prec. | Rec. | F1   |
| K    | 0.40  | 1.0  | 0.57 | --    | --   | --   |
| SA   | 0.63  | 0.45 | 0.53 | --    | --   | --   |
| ABSA | 0.77  | 0.26 | 0.39 | --    | --   | --   |
| SD 1 | 0.54  | 0.44 | 0.49 | 0.30  | 0.59 | 0.40 |
| SD 2 | 0.56  | 0.34 | 0.42 | 0.28  | 0.43 | 0.34 |
| SD 3 | 0.59  | 0.48 | 0.53 | 0.26  | 0.43 | 0.33 |
| SD 4 | 0.61  | 0.40 | 0.48 | 0.27  | 0.37 | 0.31 |
| SD 5 | 0.58  | 0.50 | 0.53 | 0.28  | 0.53 | 0.37 |
| SD 6 | 0.63  | 0.41 | 0.49 | 0.26  | 0.42 | 0.32 |
| SD 7 | 0.70  | 0.13 | 0.22 | 0.23  | 0.22 | 0.23 |
| SD 8 | 0.62  | 0.25 | 0.25 | 0.26  | 0.37 | 0.30 |
| SD 9 | 0.65  | 0.14 | 0.23 | 0.23  | 0.26 | 0.25 |

The keyword search results show that all the disinformation in the dataset explicitly mentioned one of the two putative cures, but only 40% of the data that mentions them is actual disinformation. Both SA and ABSA succeeded in narrowing down the keyword results to a pool with greater precision (with the ABSA approach nearly doubling the keyword search's precision), but again, at a cost of significantly reduced recall. The SD algorithm showed some positive results: the precision scores of all "agree" queries exceeded that of the keyword search, with some even surpassing the SA approach. The negated queries did not fare as well, showing generally worse performance than the regular claims. While the longer, more descriptive queries appear roughly to yield better precisions than the shorter queries, the trend is not found in the negative queries.

4.2.2 The "COVIDLies" Dataset

The COVIDLies dataset[21] compiles 6500 tweets that agree with, disagree with, or are unrelated to over 60 "misconceptions" about COVID-19 [99]. For each claim, most of the obtained tweets are unrelated to the claim, with only small numbers of text agreeing or disagreeing; more than any other test run so far, this dataset will required the algorithms to find relevant needles of disinformation in haystacks of otherwise irrelevant text.

Two claims from this dataset were tested. The first was "Coronavirus is genetically engineered." In the data we obtained, there were 16 tweets agreeing with this claim,

---

[21] Available at https://github.com/ucinlp/covid19-data

5 disagreeing with it, and 82 unrelated to it. The keyword search for this claim declared disinformation if the text contained either "COVID" or "coronavirus" and contained (as strings, not words) both "genetic" and "engineer." For SA, the likely sentiment is not so clear, so two tests were conducted: one in which the tweet must contain the keywords and have a negative sentiment, and a similar one but for a neutral sentiment. For ABSA, a similar difficulty exists, so again we two tests were conducted: one where any instance of either "coronavirus" or "COVID" has a negative tag, and a similar one but with a neutral tag. For both tests, the tags for the other strings can be anything. Finally, three claims were tested for the SD algorithm to agree with. The first was the given claim, "Coronavirus is genetically engineered," while the other two were the same, but obtained by replacing "coronavirus" with "COVID" and "The COVID virus" respectively. Noting that if an algorithm did not obtain any true positives, the precision, recall, and F1 are listed as 0.0, the results are shown in Table 6.

Table 6: Precision, recall, and F1 scores for a search for disinformation about COVID being genetically engineered

|        | Prec. | Rec. | F1   |
|--------|-------|------|------|
| K      | 1.0   | 0.06 | 0.12 |
| SA 1   | 0.0   | 0.0  | 0.0  |
| SA 2   | 1.0   | 0.0  | 0.12 |
| ABSA 1 | 0.0   | 0.0  | 0.0  |
| ABSA 2 | 0.0   | 0.0  | 0.0  |
| SD 1   | 0.24  | 0.56 | 0.34 |
| SD 2   | 0.28  | 0.56 | 0.38 |
| SD 3   | 0.28  | 0.56 | 0.38 |

This test highlights a weakness for the keyword-based approaches for disinformation detection: semantically similar means of expressing the same idea may be overlooked. From the keyword results here, only 1 of the 21 related tweets actually satisfied the terms of the keywords search. Unsurprisingly, since the SA and ABSA algorithms are unable to declare disinformation if the keyword condition is not satisfied, three of the four searches between them returned no true positives. Conversely, the stance detection approach is not limited by the presence or absence of particular strings, and can instead use internalized semantic understanding to assess agreement or lack thereof even when particular words may be missing in the input text. In principle, a stance detection could identify a sentence like "COVID was manufactured in a lab" as agreeing with the claim about genetic engineering, even though those two words are absent, whereas the keyword-based search categorically could not, unless an additional set of search terms containing "manufacture," "lab," etc., are used. Indeed, in this test, all three stance detection queries managed to return more than half of the text that agreed with the claim.

The second disinformation claim tested in this dataset is "coronavirus can only survive in cold temperatures," which in the data obtained, had 5 tweets agreeing with it, 2 disagreeing with it, and 35 unrelated to it. The keyword search required either "COVID" or "coronavirus" to be present and the word "cold" to be present. As with the previous test, we examined negative and neutral sentiment respectively for the SA approach, and the same respective emotions targeted at "COVID" and "coronavirus" for the ABSA approach. The two SD claims that text must agree with were the base claim itself, and the base claim with "coronavirus" replaced with "COVID." The results are given in Table 7.

Once again, only the SD algorithm was able to pick out any of the disinformation (and even then, was only able to pick out one of the 5 false sentences, with low precision). This is consonant with the idea discussed for the previous two tests: there are many ways to express the idea that COVID can only survive in the cold (e.g. saying that warm weather will kill COVID, or that COVID will go away after the winter ends), that the keyword-based searches cannot categorically detect (unless those other keywords are included in additional searches), whereas the SD algorithm at least has the possibility of understanding that only surviving in cold also means, for instance, necessarily dying in heat.

Table 7: Precision, recall, and F1 scores for a search for disinformation about COVID surviving only in the cold

|        | Prec. | Rec. | F1   |
|--------|-------|------|------|
| K      | 0.0   | 0.0  | 0.0  |
| SA 1   | 0.0   | 0.0  | 0.0  |
| SA 2   | 0.0   | 0.0  | 0.0  |
| ABSA 1 | 0.0   | 0.0  | 0.0  |
| ABSA 2 | 0.0   | 0.0  | 0.0  |
| SD 1   | 0.08  | 0.20 | 0.12 |
| SD 2   | 0.14  | 0.2  | 0.17 |

4.3 TRUTH-BASED DISINFORMATION DETECTION TESTING

The final kind of test used datasets with "true"/"false" labels, providing a concrete display of how well each approach can be used to identify false information. The tests are still somewhat imperfect, as the dataset does not identify the relevance of the false data *to the topic or claim of interest*, i.e. in principle, high scores indicating accurate identification of false information does not necessarily mean the data is false and relevant. As a consequence, for all three approaches (i.e. including stance detection), we took the presence or absence of pre-specified keywords to approximate the relevance of the text to the topic/claim of interest. Thus, the keyword search always yielded a recall score of 1.0, and the primary test of interest here was to see how well each technique can be used to improve upon the precision score of the keyword search results. Moreover, the claims/topics about which there was disinformation were not provided by the datasets themselves. Thus claims/topics to test were chosen by manual inspection of the datasets and taking claims that had reasonable numbers of related true and false tweets.

### 4.3.1 The "Countering COVID-19 Misinformation" Dataset

We used the "fake cures" subset of this dataset [100][22] to run two tests, for which we obtained approximately 2000 tweets with labels for the three classes (misinformation, legitimate information, and unrelated) in roughly equal quantity. The first test involves a claim that drinking water can make one immune to COVID-19. For the purposes of this test, a tweet was considered to be about this claim if it contained either "coronavirus" or "COVID" and the word "water," and an algorithm was considered to have identified misinformation about the claim if the algorithm declared disinformation on input text that contained those keywords and had the class label of "misinformation" in the dataset. The keyword search simply used the keywords identified above, while the SA algorithm was used for two searches, one for text which contains the keywords and is positive, and one for which the sentiment is neutral. For ABSA, four searches were used respectively: one in which "water" must have a positive tag, one in which "water" must have a neutral tag, and then two in which either "COVID" or "coronavirus" must have negative tag, and then either must have the neutral tag. For all searches, all other keywords were not required to have a specific sentiment tag. Four claims were used for SD: "Drinking water can make you immune to $x$," where $x$ is, respectively, "the coronavirus," "COVID," "COVID and the coronavirus," and "COVID or the coronavirus." The results are shown in Table 8.

As expected, the keyword search recall was a perfect 1.0 (because the keywords searched were the same keywords taken to indicate relevance to the claim of interest), but the precision score showed that only thirty percent of tweets in the dataset with those words were misinformation. The positive SA search yielded a reasonable increase in precision, though the neutral search yielded no benefits. Interestingly, the best-performing ABSA search involved using the neutral tag for water, resulting in an improved precision score over the top SA search, but a substantially reduced recall. The SD approaches appear to yield no additional benefits, with the variations between the four claims used resulting in only minor variation in performance.

*Table 8: Precision, recall, and F1 scores for searches for disinformation about two possible cures for COVID*

|  | WATER | | | GARLIC | | |
|---|---|---|---|---|---|---|
|  | Prec. | Rec. | F1 | Prec. | Rec. | F1 |
| **K** | 0.30 | 1.0 | 0.47 | 0.24 | 1.0 | 0.38 |
| **SA 1** | 0.49 | 0.58 | 0.53 | 0.48 | 0.72 | 0.57 |
| **SA 2** | 0.27 | 0.13 | 0.17 | 0.12 | 0.09 | 0.10 |
| **ABSA 1** | 0.3 | 0.01 | 0.2 | 0.34 | 0.12 | 0.17 |
| **ABSA 2** | 0.53 | 0.24 | 0.33 | 0.19 | 0.12 | 0.15 |
| **ABSA 3** | 0.50 | 0.05 | 0.10 | 0.12 | 0.02 | 0.03 |
| **ABSA 4** | 0.30 | 0.01 | 0.02 | 0.25 | 0.02 | 0.03 |
| **SD 1** | 0.23 | 0.66 | 0.34 | 0.11 | 0.84 | 0.19 |
| **SD 2** | 0.27 | 0.52 | 0.36 | 0.13 | 0.82 | 0.22 |
| **SD 3** | 0.23 | 0.63 | 0.34 | 0.11 | 0.86 | 0.19 |
| **SD 4** | 0.23 | 0.58 | 0.33 | 0.11 | 0.85 | 0.19 |

The second test involved the idea that garlic can be used to prevent infection by the coronavirus. The same SA, ABSA, and SD tests run for the water disinformation test were run for the garlic claim as well (i.e. replacing "water" with "garlic"), with the minor caveat that the specific base SD claim used was "Garlic prevents infection by $x$". The results, also shown in Table 8, are largely similar to the first test. The keyword search indicated that 24 percent of the discussion about garlic is misinformation, and the positive SA search doubled the precision of the returned results. The ABSA search looking for positive discussion of garlic yielded an improved precision as well, but not as high as the SA results, and with a much lower recall. The SD approach had a higher recall than the sentiment approaches, but a very low precision, ending up uniformly worse than even the keyword search.

### 4.3.2 The "CMU-MisCov19" Dataset

The final test dataset was derived from a dataset of tweets labelled not merely with "true" or "false," but with finer-grained labels distinguishing various kinds of disinformation [101, 102].[23] For the purposes of this paper, we considered any text in the dataset with the labels "Conspiracy," "Fake Cure," "Fake Treatment,"

---

[22] Available at sites.google.com/view/counter-covid19-misinformation

[23] Available at zenodo.org/record/4024154#.YMupJulKj_Q

"False Fact or Prevention," and "False Public Health Response," to count as "false;" all others were counted simply as not being false (but not necessarily true, e.g. as with the class "Irrelevant"), and concerned ourselves with testing the ability of the stance and sentiment algorithms to pick out the text that counts as false.

Manually perusing the data, two prominent narratives emerged. One pertains to the possibility that the coronavirus was deliberately designed as a bioweapon. The keywords search for this claim was "COVID" or "coronavirus" and "bioweapon." The sentiments used were negative and neutral, respectively. For the ABSA search, four tests were used: the first two required negative and neutral tags, respectively, to appear as the tag on "bioweapon," while the latter two required the same respective sentiments to be the tags for "COVID" or "coronavirus." Two queries for stance detection were used, both of the form "$x$ is a bioweapon," with $x$ being "COVID" and "coronavirus" respectively. The results are shown in Table 9.

In this case, the keyword search was more valuable by itself than any of the machine learning algorithms. In this dataset, most of the discussion about COVID being a bioweapon is disinformation/false (76% according to the keyword search), and so there is little room for the algorithms to improve upon the basic keyword search. The only search that does, the aspect-based search for "neutral" tags on "bioweapon," does so only with a drastically reduced recall.

Table 9: Precision, recall, and F1 scores for searches for disinformation about COVID being a bioweapon and being caused by 5G

|  | BIOWEAPON | | | 5G | | |
| --- | --- | --- | --- | --- | --- | --- |
|  | Prec. | Rec. | F1 | Prec. | Rec. | F1 |
| K | 0.76 | 1.0 | 0.87 | 0.3 | 1.0 | 0.47 |
| SA 1 | 0.69 | 0.50 | 0.58 | 0.27 | 0.68 | 0.39 |
| SA 2 | 0.76 | 0.22 | 0.34 | -- | -- | -- |
| ABSA 1 | 0.71 | 0.04 | 0.08 | 0.15 | 0.13 | 0.14 |
| ABSA 2 | 1.0 | 0.02 | 0.03 | -- | -- | -- |
| ABSA 3 | 0.50 | 0.04 | 0.08 | -- | -- | -- |
| ABSA 4 | 0.75 | 0.08 | 0.14 | -- | -- | -- |
| SD 1 | 0.11 | 0.57 | 0.19 | 0.02 | 0.48 | 0.05 |
| SD 2 | 0.11 | 0.57 | 0.18 | 0.03 | 0.46 | 0.05 |

The second test has to do with 5G being the cause of the coronavirus. The keyword search required "COVID" or "coronavirus" and "5G," and the SA search required a negative sentiment. Given that the belief that 5G causes coronavirus is likely to engender negative feelings about 5G itself, the aspect-based search looked for sentences that have negative feelings about 5G itself, rather than towards the "COVID" or "Coronavirus" words used in the previous tests. The results are also shown in Table 9.

Here, the results are somewhat surprising. Despite only thirty percent of mentions of coronavirus/COVID and 5G in the dataset being disinformation, none of the algorithms were particularly successful at narrowing down results to a more concentrated pool of disinformation. Here, the explanation of there not being much more room to improve is not available; indeed, it may simply be that these algorithms are not well-suited to pick out disinformation about 5G causing coronavirus, whether because such discussion is not typically sentimental or because the algorithms themselves perform worse on out-of-distribution data.

## 5   DISCUSSION AND CONCLUSION

The title of this paper poses a question, asking whether stance, sentiment (or something else) might be best-suited for use in human-in-the-loop disinformation detection. The results of the tests run herein, however, are not sufficient to answer it conclusively, but do illustrate the need to ask and study the question in the first place. As discussed in Section 4, the absence of an ideal dataset for this question means that all tests conducted were indirect in one way or another, and issues associated with small numbers of positive examples and the possibility of unrepresentative datasets may also limit the generality of the results obtained here. Nevertheless, the results that were obtained do hint at possibilities that could be impactful for how analysts use machine learning to identify disinformation and thus provide reasonable motivation for further study.

The most noteworthy point suggested by our data is the apparent utility of SA for disinformation detection, despite the fact that this approach has traditionally been overlooked in the literature on disinformation detection. Recall that all three methods (SA, ABSA, and SD) all use nearly identical (and near-state-of-the-art) architectures (Section 3.2), and were trained on datasets of equal size (Section 3.3); the stance detection validation dataset test (Section 3.4) also showed that the SD algorithm is capable of achieving near-state-of-the-art performance on one of the most popular stance-based disinformation detection datasets in the literature. Despite all this, the SA algorithms routinely outperformed the SD algorithms on precision[24] and F1 scores, even in cases where the test

---

[24] Given the volume of results that even targeted keyword searches for data are likely to return, we argue that precision is

dataset was originally designed for a stance detection task: see e.g. Sections 4.1.1 and 4.2.1. This is possibly explained (at least in part) by the fact that many disinformation narratives do seem to have a sentimental component to them: believing that some thing is a cure for coronavirus, whether it be water, garlic (Section 4.3.1), or (hydroxy)chloroquine (Section 4.2.1), suggests that one would also feel positively about those things, while thinking that something is a cause of (or exacerbates) coronavirus, like 5G (Section 4.3.2), might reasonably result in negative sentiment towards those things. In all of these instances, SD provided no appreciable improvement over SA or ABSA in identifying disinformation, with SA and/or ABSA providing clear advantages in some (but not all) instances. The only circumstance in which the SD algorithm seemed to provide a clear advantage was in Section 4.2.2, where we tested on some narratives that either had no obvious sentimental component and where the narrative was not well summarized into a single set of keywords (e.g. "The coronavirus can only survive in cold weather").

In addition to considering the relative merits of each technique, it is worth examining the benefits of using these techniques for disinformation detection at all. To the best of our knowledge, this paper is the only work that addresses disinformation detection for the specific circumstances outlined in Section 1, but "the best available" methods are not the same as "good" methods. Consider the following: the three algorithms used near state-of-the-art architectures and relatively large, balanced (where possible), and diverse datasets for training. When tested on task specific datasets (Section 4.1), they showed some drop in performance compared to the validation datasets (Section 3.4), but not an unreasonable amount given the difference in data sources. But yet, in some cases, when tested for disinformation detection ability, the algorithms where unable to produce results much better than what a keyword search could do (see e.g. Section 4.3.2). Perhaps this is a sign that these tasks are not all that well suited to being repurposed for use in disinformation detection. It may be that the best use of the techniques examined in this paper for disinformation detection is as a stop gap measure: as new disinformation narratives arise on short-term scales, these techniques may be the only methods that can be used to detect the new disinformation on short notice (see Section 1). However, if the disinformation narratives are expected to last (as many have with COVID-19) it may be worthwhile to take the time to develop narrative-specific disinformation detection methods that were dismissed in Section 1 for taking too long to develop (in general), in order to obtain better performance.

The discussion in this section has hinted at a number of productive next steps for research in this area that are worth explicating in further detail. First, sentiment analysis methods (both traditional and aspect-based) should be studied further for use in disinformation detection, whether partially or fully-automated. Additionally, greater effort should be spent in characterizing when stance is better used than sentiment and vice versa, i.e. to establish whether or not it really is the case that when searching for disinformation narratives with strong sentimental components, one should use keywords plus sentiment, while when searching for more complex narratives with little sentiment or not easily summarized with a few keywords, one should use stance detection. Stance detection research for disinformation detection must also be continued, with a focus on ensuring that the methods developed are more robust to novel domains (hence the focus on using COVID-related testing data here), and determining what auxiliary sentences are best used in this application – no discernable patterns emerged from the tests here. The possibility of using NLI or QA (see Section 2.5) for disinformation detection is also worth pursuing, given that the SA, ABSA, SD algorithms used here did not always produce desired results. Finally, more appropriate and representative datasets should be developed and tested for this kind of research: as previously discussed, such a dataset would have labels indicating truth or falsity but also labels indicating broadly what the text is about (so the test can distinguish between relevant and irrelevant disinformation), but would also contain representative proportions of disinformation to non-disinformation, to facilitate testing of just how well candidate algorithms separate signal from noise. Such a dataset would be important for ensuring that candidate algorithms are not merely successful at improving bespoke metrics on curated academic dataset, but are really substantially capable of helping humans succeed in the practical task of identifying disinformation relevant to their interests.

**REFERENCES**

---

the more important metric (between precision and recall) for understanding efficacy at disinformation detection: the difficulty is not in ensuring that the data contains needles of relevant disinformation, so much as it is about locating those needles in the haystacks of irrelevant information and disinformation.